\documentclass{article}

\usepackage[preprint]{neurips_2019}

\usepackage[utf8]{inputenc} 
\usepackage[T1]{fontenc}    
\usepackage{url}            
\usepackage{booktabs}       
\usepackage{amsfonts}       
\usepackage{nicefrac}       
\usepackage{microtype}      
\usepackage{svg}
\usepackage[font=footnotesize,labelfont=bf]{caption}
\usepackage{todonotes}

\usepackage{color}

\usepackage{etoolbox}

\newbool{draftmode}
\boolfalse{draftmode}

\usepackage{afterpage}

\usepackage{algorithm}

\usepackage{amsmath}

\usepackage{amssymb}

\usepackage{amsthm}

\usepackage{array}

\usepackage{booktabs}

\usepackage{caption}
\usepackage{enumitem}

\usepackage{fixme}

\usepackage{graphicx}

\usepackage{mathtools}

\usepackage{microtype}

\usepackage{multirow}

\usepackage{setspace}

\usepackage{subfig}

\usepackage{tabu}

\usepackage{tabularx}

\usepackage{tabulary}

\usepackage{titlesec}

\usepackage{xfrac}

\usepackage[
]{algorithmic}

\setcitestyle{numbers,square,compresss,sort}

\DeclareMathOperator*{\argmax}{argmax}
\DeclareMathOperator*{\argsort}{argsort}

\title{Model-based Lookahead Reinforcement Learning}

\author{
    Zhang-Wei Hong \\
    National Tsing Hua University \\
    \texttt{williamd4112@gapp.nthu.edu.tw} 
    \And
    Joni Pajarinen \\
    TU Darmstadt \\
    \texttt{joni@robot-learning.de} 
    \And
    Jan Peters \\
    TU Darmstadt and Max Planck Institute for Intelligent Systems \\
    \texttt{mail@jan-peters.net}
}

\begin{document}

\maketitle
\vspace{-5ex}
\begin{abstract}


{
Model-based Reinforcement Learning (MBRL) allows data-efficient learning which is required in real world applications such as robotics. However, despite the impressive data-efficiency, MBRL does not achieve the final performance of state-of-the-art Model-free Reinforcement Learning (MFRL) methods. We
leverage the strengths of both realms and propose an approach that 
obtains high performance with a small amount of data. In particular, we combine MFRL and Model Predictive Control (MPC). While MFRL's strength in exploration allows us to train a better forward dynamics model for MPC, MPC improves the performance of the MFRL policy by sampling-based planning. The experimental results in standard continuous control benchmarks show that our approach can achieve MFRL`s level of performance while being as data-efficient as MBRL.}
\end{abstract}

\section{Introduction}
\label{sec::intro}
Model-free Reinforcement Learning (MFRL) has succeeded in several domains, including video game playing~\cite{mnih2015human, mnih2016a3c} and robot control~\cite{lillicrap2016ddpg, schulman2015trust}. However, high sample complexity prevents applying MFRL to most complex real world applications: MFRL directly optimizes the agent's policy from interactions with the environment usually requiring millions of data samples~\cite{mnih2015human,mnih2016a3c,lillicrap2016ddpg}. 

Contrary to MFRL, model-based reinforcement learning (MBRL) is typically more data-efficient. However, when the perfect model is inaccessible, a forward dynamics model of the environment must be approximated~\cite{deisenroth2011pilco, kurutach2018modelensemble,nagabandi2017neural,kamthe2018gpmpc,chua2018deep}. Despite the learning of the forward dynamics model, MBRL often requires less interactions with the environment compared to learning a policy directly. Nevertheless, a major disadvantage of MBRL is that, due to model approximation errors~\cite{nagabandi2017neural}, MBRL commonly cannot achieve the performance of MFRL at convergence.

Prior work takes advantage of both MFRL and MBRL to some extent. \citet{levine2013guided} show that generating training samples for MFRL by model-based trajectory optimization may increase the performance of MFRL, while~\citet{gu2016continuous} find in contrast that insufficient exploration of MBRL may impair the performance of the resultant policy and ~\citet{nagabandi2017neural} show that naive exploration for training forward dynamics models prevents accuracy of the acquired forward dynamics model. Apart from generating samples using MBRL, \citet{silver2016mastering,tamar2016value,oh2017value,lowrey2018plan} combine MFRL with online model-based planning (i.e. MBRL) to improve the performance of the reinforcement learning (RL) agent, but either is restricted in discrete state and action spaces, relies on the assumption that the state space has 2-D structure, or assumes a perfect forward dynamics model. In sum, despite the notable performance of the conjunction of MFRL with model-based planning, the strong assumptions prohibit applications on more complex tasks.

We propose a novel framework that unifies MFRL and MBRL through Model Predictive Control (MPC). Our approach leverages the merits of MFRL and MBRL at both training and testing time. At training time, we utilize the exploratory policy of MFRL to collect more diverse training samples in the environment than MBRL~\cite{gu2016continuous,nagabandi2017neural} and thus prevent the impacts of insufficient exploration on policies and forward dynamics models. Next, at testing time, we combine sampling-based MPC (a model-based online planning approach) with MFRL to further increase the performance beyond training. Different from the contemporary works~\cite{silver2016mastering,lowrey2018plan}, our approach does not rely on any assumption of perfect forward dynamics models or state/action space. We use an approximated forward dynamics model that can be applied in arbitrary state/action spaces. Furthermore, we jointly leverage the value function and the policy to yield remarkable MPC planning performance and data-efficiency. Finally, we propose a soft-greedy approach that improves action selection in planning. To our best knowledge, we are the first to combine MFRL and MPC, on the least assumptions of state/action space.

We demonstrate the effectiveness of our approach on well-known challenging continuous control tasks in MuJoCo~\cite{todorov2012mujoco}. The experimental results show that our approach leads to better performance in less data than that of independently using MFRL or MPC, particularly in complex tasks, and thus confirm that our approach of combining MBRL and MFRL is effective. Furthermore, the results show that our approach favors the quality of forward dynamics model training. In addition, we provide empirical analysis to verify the limitations of contemporary approaches.

The contributions of this paper are the following:
\begin{itemize}
    \item We show that using an MFRL policy to collect training data improves the quality of the trained forward dynamics model.
    \item We show that using an MFRL policy can enhance MPC` s performance.
    \item We show advantages and limitations of using value function in MPC.
    \item We provide a comprehensive evaluation of each design decision for combining MPC and MFRL.
\end{itemize}

The remaining parts of this paper are organized as follows. Section~\ref{sec::bg} introduces the background. Section~\ref{sec::approach} describes our new approach. Section~\ref{sec::exp} presents and analyzes the experimental results. Finally, Section~\ref{sec::conclusion} concludes the paper.

\section{Background}
\label{sec::bg}
In order to elaborate the motivation and the implementation of the proposed method, this section starts with an introduction to Reinforcement Learning (RL)~\cite{sutton1998introduction}, then explains MFRL, and finally discusses recently proposed MPC approaches for MBRL.
\subsection{Reinforcement Learning (RL)} 
\label{subsec::rl}
In this paper, we study a standard deterministic discrete time RL problem, consisting of a 4-tuple $(\mathcal{S}, \mathcal{A}, R, f)$. $\mathcal{S}$ denotes the state and $\mathcal{A}$ the action space. $R:\mathcal{S} \times \mathcal{A} \mapsto \mathbb{R}$ is a task-specific reward function encoding the task objective. $f:\mathcal{S} \times \mathcal{A} \mapsto \mathcal{S}$ is the true forward dynamics model. At each time step $t$, an RL agent perceives the current state $\boldsymbol{s}_t \in \mathcal{S}$, takes the control action $\boldsymbol{a}_t \in \mathcal{A}$ and observes the next state $\boldsymbol{s}_{t+1} = f(\boldsymbol{s}_t, \boldsymbol{a}_t)$ and the immediate reward $R(\boldsymbol{s_t}, \boldsymbol{a}_t)$. The training objective of RL is to search for an optimal controller that selects $\boldsymbol{a}_t$ for each $\boldsymbol{s}_t$ such that the expected return $G_t = \sum_{i=t}^{T} \gamma^{i-t} R(\boldsymbol{s}_i, \boldsymbol{a}_i)$ is maximized, where $T$ represents the horizon of the task and $\gamma \in (0,1)$ denotes the constant discount factor.  Note that while this paper adheres to deterministic cases for simplicity our framework can be easily extended to a stochastic formulation. 

\subsection{Model-free Reinforcement Learning (MFRL)}
\label{subsec::mfrl}
We discuss both the training and evaluation phases in MFRL. First, using data $(\boldsymbol{s}_t, \boldsymbol{a}_t, \boldsymbol{s}_{t+1})$ collected by an exploratory policy $\pi^\prime$, the training phase learns a possibly stochastic control policy $\pi_{\boldsymbol{\theta^\pi}}: \mathcal{S} \times \mathcal{A} \mapsto [0, 1]$ that infers $\boldsymbol{a}_t$ which maximizes $G_t$ for $\boldsymbol{s}_t$, and a value function $V_{\boldsymbol{\theta^V}}: \mathcal{S} \mapsto \mathbb{R}$ estimating expected returns for $\boldsymbol{s_t}$. $\boldsymbol{\theta^\pi}$ and $\boldsymbol{\theta^V}$ denote parameters. In this paper, we adhere to an appealing MFRL approach, policy gradient~\cite{sutton2000policy} which updates $\boldsymbol{\theta^\pi}$ along the direction $\nabla_{\boldsymbol{\theta^\pi}}  \text{log}(\pi_{\boldsymbol{\theta^\pi}}(\boldsymbol{a}_t|\boldsymbol{s}_t)) (G_t - b(\boldsymbol{s}_t))$. $b(\boldsymbol{s}_t)$ is for variance reduction~\cite{williams1992simple}. A common choice for $b(\boldsymbol{s}_t)$ is the value function estimate $V_{\boldsymbol{\theta}^V}(\boldsymbol{s}) \approx \mathbb{E}_{\boldsymbol{s}_{t+1:T}, \boldsymbol{a}_{t:T} \sim \pi_{\boldsymbol{\theta^\pi}}}\big[G_t | \boldsymbol{s_t} = \boldsymbol{s} \big]$. $\boldsymbol{\boldsymbol{\theta}^V}$ can be obtained by various value function approximation approaches~\cite{schulman2015high,sutton1998introduction,bertsekas2005dynamic}. Finally, the evaluation phase simply samples control actions $\boldsymbol{a}_t$ using $\pi_{\boldsymbol{\theta}^\pi}$: $\boldsymbol{a}_t \sim \pi_{\boldsymbol{\theta}^\pi}(\boldsymbol{a}|\boldsymbol{s}_t)$.

\subsection{Model Predictive Control for Model-based Reinforcement Learning (MPC-MBRL)}
\label{subsec::mpc}
MPC has long been prevalent in robotic control~\cite{garcia1989model,tassa2014control} and recently been applied to MBRL~\cite{nagabandi2017neural,kamthe2018gpmpc,chua2018deep,williams2017information}. MPC-MBRL simply trains a forward dynamics model to plan the control action $\boldsymbol{a}_t$ at each time step during evaluation.

The training phase trains an approximated forward dynamics model $f_{\boldsymbol{\theta^f}}$ since the real forward dynamics model $f$ is usually inaccessible~\cite{nagabandi2017neural,kamthe2018gpmpc}.
MPC-MBRL collects an initial dataset $\mathcal{D}$ consisting of a series of transitions $(\boldsymbol{s}_t, \boldsymbol{a}_t, \boldsymbol{s}_{t+1})$ by using uniformly random exploration in the environment:  $\boldsymbol{a}_t \sim \mathcal{U}(\mathcal{A})$, then optimizing $f_{\boldsymbol{\theta^f}}$ by minimizing the following loss function:
\begin{equation}
\label{eq::fwd_dynamic_loss}
    L_{{f_{\boldsymbol{\theta}^f}}}(\boldsymbol{\theta}_f) = \frac{1}{\lVert {B} \rVert} \sum_{\boldsymbol{s}_t, \boldsymbol{a}_t, \boldsymbol{s}_{t+1} \in {B}}{{\lVert \boldsymbol{s}_{t+1} - {f_{\boldsymbol{\theta}^f}}(\boldsymbol{s}_t, \boldsymbol{a}_t) \rVert}^2},
\end{equation}
where $B$ is a batch of transitions $(\boldsymbol{s}_t, \boldsymbol{a}_t, \boldsymbol{s}_{t+1})$ sampled from $\mathcal{D}$. Next, MPC-MBRL appends data from online execution of MPC-planning to $\mathcal{D}$: $\boldsymbol{a}_t = \text{MPC}(\boldsymbol{s}_t)$~\cite{nagabandi2017neural,chua2018deep}, and trains the forward dynamics model $f_{\boldsymbol{\theta^f}}$ according to Eq.~\ref{eq::fwd_dynamic_loss} again. 

In the evaluation phase, MPC-MBRL computes at each time step the control action $\boldsymbol{a}_t$ for the current state $\boldsymbol{s}_t$, with model-based planning consisting of three stages: trajectory sampling, trajectory evaluation, and action selection.
Trajectory sampling stage simulates a set of trajectories: $\Phi = \{\hat{\tau}^n = [\boldsymbol{\hat{s}}^n_1, \boldsymbol{\hat{a}}^n_1, \boldsymbol{\hat{s}}^n_2, \cdots, \boldsymbol{\hat{s}}^n_{H+1}]\}^{N}_{n=1}$, where $N$ denotes the number of simulated trajectories, $n$ denotes the index of a trajectory within $\Phi$, and $H$ indicates the planning horizon. $\tau^n$ can be obtained by sequentially applying
\begin{equation}
    \begin{aligned}
    \label{eq::traj_sampling}
        \boldsymbol{\hat{s}}^n_1 &= \boldsymbol{s}_t; ~  \boldsymbol{\hat{s}}^n_{h+1} = f_{\boldsymbol{\theta^f}}(\boldsymbol{\hat{s}}^n_h, \boldsymbol{\hat{a}}^n_h); ~ \boldsymbol{\hat{a}}^n_h \sim \mathcal{Z},
    \end{aligned}    
\end{equation}
where $\boldsymbol{\hat{s}}^n_h$ denotes the simulated state at planning step $h$ within $\hat{\tau}^n$, $\boldsymbol{\hat{a}}^n_h$ is the action applied to $\boldsymbol{\hat{s}}^n_h$, and $\mathcal{Z}$ denotes a given action distribution.
Next, trajectory evaluation stage evaluates each trajectory $\hat{\tau}^n$ with the task-specific reward function $R$ and the terminal reward function $R_{\phi}$:
\begin{equation}
    \label{eq::traj_eval}
     \hat{G}(\boldsymbol{\hat{s}}^n_{1:H+1}, \boldsymbol{\hat{a}}^n_{1:H}) =  {\sum^{H}_{h = 1}{\gamma^{h - 1}R(\boldsymbol{\hat{s}}^n_h, \boldsymbol{\hat{a}}^n_h})} + \gamma^{H} R_{\phi}(\boldsymbol{\hat{s}}^n_{H}),
\end{equation}
where $\hat{G}(\boldsymbol{\hat{s}}^n_{1:H+1}, \boldsymbol{\hat{a}}^n_{1:H})$ denotes the simulated accumulated rewards associated with $\hat{\tau}^n$, and $\boldsymbol{\hat{s}}^n_{1:H+1}$ and $\boldsymbol{\hat{a}}^n_{1:H}$ denote the state and action sequences extracted from $\hat{\tau}^n$ respectively. Although $R_{\phi}$ is typically ignored (i.e.\ $R_{\phi}(.) = 0$), we use $R_{\phi}$ to illustrate our approach in Section~\ref{sec::approach}.
Finally, action selection stage selects the first control action $\boldsymbol{a}^*_t$ of the action sequence $\boldsymbol{a}^*_{t:t+H}$ which yielded highest value w.r.t.~Eq.~\ref{eq::traj_eval}:
\begin{equation}
    \label{eq::act_select}
    \boldsymbol{a}^*_{t:t+H} =  \argmax_{\boldsymbol{\hat{a}}^n_{1:H}}{{\hat{G}(\boldsymbol{\hat{s}}^n_{1:H+1}, \boldsymbol{\hat{a}}^n_{1:H})}}.
\end{equation}


\section{The approach: Model Predictive Control with Model-free Reinforcement Learning (MPC-MFRL)}
\label{sec::approach}
Typical MFRL algorithms and MBRL methods do not fully utilize the beneficial information that can be extracted from the environment. MFRL ignores the information encapsulated in the forward dynamics of the environment, while MPC neglects the utility of policies and value functions. Contrary to both, our approach jointly leverages forward dynamics, policies, and value functions, and therefore is more likely to have better performance in the case of dearth of data.
Section~\ref{subsec::fwd_dynamic} details the proposed training method and Section~\ref{subsec::mpc_act_select} the proposed online hybrid MPC-MFRL approach.


\begin{figure*}[t]
    \centering
    \begin{minipage}[b]{.95\textwidth}
    	\centering
    	\includegraphics[width=\linewidth]{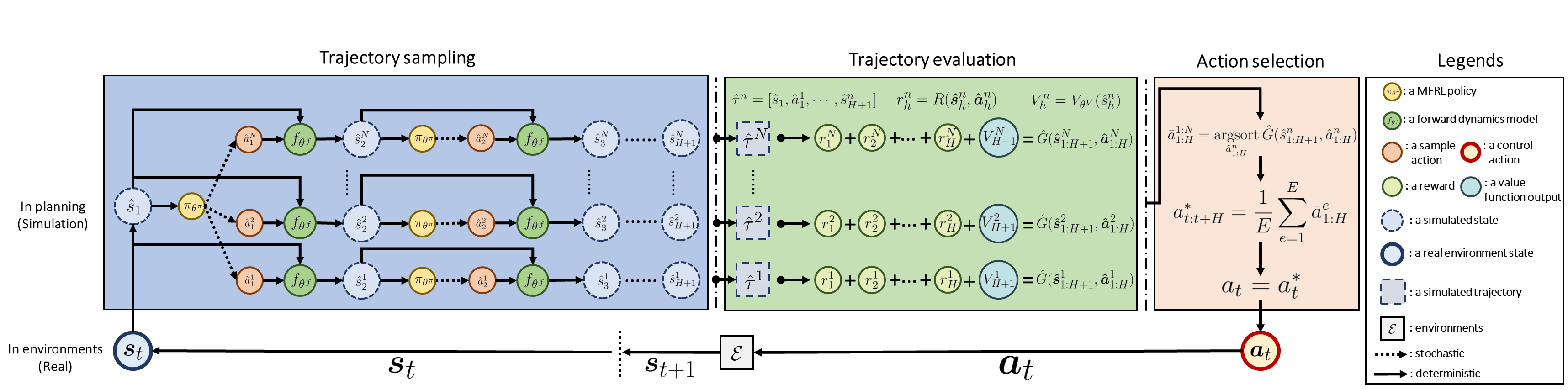}
    	\caption{Overview of MPC-MFRL at evaluation time: In state $\boldsymbol{s}_t$, MPC-MFRL samples trajectories using an MFRL policy, evaluates sampled trajectories by an MFRL value function, and then chooses an action $\boldsymbol{a}_t$ based on Eq.~\ref{eq::act_select}. The environment transitions to $\boldsymbol{s}_{t+1}$ and the process starts from the beginning. The upper row illustrates planning in simulation. The lower row depicts interaction with the real environment.}
    	\label{fig::mpmfrl}
    \end{minipage}
\end{figure*}

\subsection{The training phase: learning a policy, value function, and a forward dynamics model}
\label{subsec::fwd_dynamic}
We jointly train the control policy $\pi_{\boldsymbol{\theta}^\pi}$, the value function $V_{\boldsymbol{\theta}^V}$, and the forward dynamics model $f_{\boldsymbol{\theta}^{f}}$ using the same data at the same time. The training phase iteratively executes the following three steps until convergence. Firstly, we gather a trajectory $\tau_{\pi^\prime} = [\boldsymbol{s}_1, \boldsymbol{a}_1, \boldsymbol{s}_{2}, \cdots, \boldsymbol{s}_{T+1}]$ using the exploratory policy $\pi^\prime$ (if using on-policy RL, $\pi^\prime$ must be set as ${\boldsymbol{\theta}^\pi}$). $\pi^\prime$ collects data by interacting with the environment: $\boldsymbol{a}_t \sim \pi^\prime(\boldsymbol{a}|\boldsymbol{s_t})$ (as Section~\ref{subsec::rl} describes). Secondly, we update ${\boldsymbol{\theta}^\pi}$ and ${\boldsymbol{\theta}^V}$ by $\tau_{\pi^\prime}$, as Section~\ref{subsec::rl} describes. In this paper, we use Trust Region Policy Optimization (TRPO)~\cite{schulman2015trust}, a trust region policy gradient method, for training $\pi_{\theta^\pi}$ since TRPO has been successful in several domains and is a theoretic sounded policy gradient algorithm, but other MFRL methods~\cite{lillicrap2016ddpg,schulman2017proximal} could be used instead. Finally, we append $\tau_{\pi^\prime}$ to the training dataset $\mathcal{D}$: $\mathcal{D} \leftarrow \mathcal{D} \cup \{(\boldsymbol{s}_t, \boldsymbol{a}_t, \boldsymbol{s}_{t+1})  \}^T_{t=1}$ ($\mathcal{D}$ is initialized as $\emptyset$), and optimizes the forward dynamics model $f_{\boldsymbol{\theta}_f}$ by minimizing the loss function defined in Eq.~\ref{eq::fwd_dynamic_loss} with batches $B$ sampled in $\mathcal{D}$. Algorithm 5 in supplementary material details the training scheme. Note that though this paper focuses on training a deterministic forward dynamics model, probabilistic models~\cite{chua2018deep} or non-parametric models~\cite{kamthe2018gpmpc} can be easily applied in our approach. 

Jointly training $\pi_{\boldsymbol{\theta}^\pi}$, $V_{\boldsymbol{\theta}^V}$, and $f_{\boldsymbol{\theta}^f}$ poses the following advantages. First and foremost, $\pi^\prime$ can collect more extensive data for the forward dynamics model than does uniform random exploration and MPC on-policy data aggregation~\cite{nagabandi2017neural}. Optimized to maximize reward, $\pi_{\boldsymbol{\theta}^\pi}$ can collect successful experiences which uniform random exploration cannot. Also, MPC simply exploits rewards while an MFRL exploratory policy balances exploitation and exploration and thereby guarantees data diversity. Second, interacting with an environment using $\pi^\prime$ requires less computation time than collecting the training dataset using MPC-planning~\cite{nagabandi2017neural,chua2018deep}. Finally, our joint training procedure maximize the utility of each interaction with the environment, comparing to simply learning a policy (MFRL) or a forward dynamics model (MBRL).


  

\subsection{The evaluation phase: planning the control action using the policy, the value function, and the forward dynamics model}
\label{subsec::mpc_act_select}
Our approach at the evaluation time is built on the top of MPC framework defined in Section~\ref{subsec::mpc} and improves each stage by MFRL. We use MFRL`s control policy for trajectory sampling, MFRL`s value function for trajectory evaluation, and soft-greedy approach for action selection. Fig.~\ref{fig::mpmfrl} illustrates the planning process and Algorithm~\ref{alg::mbl_act} summarizes the approach. Next, we discuss how our method improves trajectory sampling, trajectory evaluation, and action selection. 

\paragraph{Trajectory sampling.} ~
By replacing the action distribution $\mathcal{Z}$ in Eq.~\ref{eq::traj_sampling} with the MFRL control policy $\pi_{\boldsymbol{\theta}^\pi}$, our method can more efficiently sample trajectories of high value than uniform random sampling~\cite{richards2005robust} and Cross Entropy Method (CEM)~\cite{rubinstein1999cross}. We sample as follows:
\begin{equation}
    \begin{aligned}
    \label{eq::new_traj_sampling}
        \boldsymbol{\hat{s}}^n_1 &= \boldsymbol{s}_t; ~ \boldsymbol{\hat{s}}^n_{h+1} = f_{\boldsymbol{\theta}^f}(\boldsymbol{\hat{s}}^n_h, \boldsymbol{\hat{a}}^n_h); ~ \boldsymbol{\hat{a}}^n_h \sim \pi_{\boldsymbol{\theta}^\pi}(\boldsymbol{a}|\boldsymbol{\hat{s}}^n_h).
    \end{aligned}    
\end{equation}
The MFRL control policy $\pi_{\boldsymbol{\theta}^\pi}$ improves trajectory sampling in the following aspects.
To begin with, $\pi_{\boldsymbol{\theta}^\pi}$ can readily result in high-value trajectories since $\pi_{\boldsymbol{\theta}^\pi}$ is trained to maximize the expected return $G_t$. Additionally, even before convergence $\pi_{\boldsymbol{\theta}^\pi}$ can restrain the search space in trajectory sampling and thus allows trajectories of high value to be sampled more likely than in uniform sampling~\cite{richards2005robust}. 
Moreover, in contrast to CEM~\cite{rubinstein1999cross} which uses simulated data for optimization,
our method is not susceptible to forward dynamics approximation error since $\pi_{\boldsymbol{\theta}^\pi}$ is trained with real experience. 
Finally, our method does not require costly computations since our method does not rely on online iterative optimization like CEM~\cite{rubinstein1999cross}.

\paragraph{Trajectory evaluation.} ~
Similar to ~\citet{lowrey2018plan}, our approach substitutes the terminal reward function $R_{\phi}$ (Eq.~\ref{eq::traj_eval}) with the MFRL value function $V_{\boldsymbol{\theta}^V}$ to resolve the shortsighted planning problem mentioned in Section~\ref{subsec::mpc}. Trajectory evaluation becomes:
\begin{equation}
    \label{eq::new_traj_eval}
     \hat{G}(\boldsymbol{\hat{s}}^n_{1:H+1}, \boldsymbol{\hat{a}}^n_{1:H}) =  {\sum^{H}_{h = 1}{\gamma^{h - 1}R(\boldsymbol{\hat{s}}^n_h, \boldsymbol{\hat{a}}^n_h)}} + \gamma^{H} V_{\boldsymbol{\theta}^V}(\boldsymbol{\hat{s}}^n_{H}).
\end{equation}
The MFRL value function $V_{\boldsymbol{\theta}^V}$ in Eq.\ref{eq::new_traj_eval} estimates the expected return of a given state (see Section~\ref{subsec::mfrl}) and therefore prevents shortsighted planning even with a short planning horizon $H$. More importantly, planning with a short horizon avoids compounding errors in simulation of long horizon and saves computation time as well. 

Different from ~\citet{lowrey2018plan}, we use an approximated dynamics model rather than a perfect one. The assumption of a perfect dynamics model is unrealistic in most cases, especially on complex tasks, where the perfect dynamics model is inaccessible. In addition, estimating the expected return of a simulated state could be risky since the value function trained with real experience is unlikely to be accurate in the states absent in the training data. Moreover, jointly using value function estimation and a plain random action sampling in simulation of MPC-planning could magnify this problem since those unconstrained or weakly constrained action sampling may induce lots of unreachable states (e.g. states that violate physical constraints) where the value function cannot estimate accurately on.

However, our approach can alleviate the above problem by sampling actions with an MFRL policy in trajectory sampling. MFRL policies can prevent agents from performing actions that could lead to unreachable states since MFRL polices are trained to maximize to expected reward of the task and thus are less likely to perform those infeasible actions.




\paragraph{Action selection.} ~ If we simply take the control action as the sampled action that yielded the max expected return (Eq.~\ref{eq::act_select}), the agent may be overly optimistic to the simulated results and is likely to impair the performance. Thus, we propose a soft-greedy approach to alleviate the impact of approximation errors in the forward dynamics model $f_{\boldsymbol{\theta}^f}$. Our soft-greedy approach takes the control action $\boldsymbol{a}_t$ as the average over the $E$ best action sequences w.r.t.~$\hat{G}(\boldsymbol{\hat{s}}^n_{1:H+1}, \boldsymbol{\hat{a}}^n_{1:H})$. Formally, we describes as the follows:
\begin{equation}
    \label{eq::new_act_select}
    \begin{split}
            \bar{\boldsymbol{a}}^{1:N}_{1:H} &=  \argsort_{\boldsymbol{\hat{a}}^n_{1:H}}{{\hat{G}(\boldsymbol{\hat{s}}^n_{1:H+1}, \boldsymbol{\hat{a}}^n_{1:H})}}. \\
            \boldsymbol{a}^*_{t:t+H} &= \frac{1}{E} \sum_{e=1}^{E}\bar{\boldsymbol{a}}^{e}_{1:H}, \\
    \end{split}
\end{equation}
where $\argsort$ sorts all action sequences $\boldsymbol{\hat{a}}^{1:N}_{1:H}$ according to $\hat{G}(\boldsymbol{\hat{s}}^n_{1:H+1}, \boldsymbol{\hat{a}}^n_{1:H})$ in descending order.

Our proposed soft-greedy action selection approach can prevent biasing toward the best action obtained from imperfect simulation. Averaging has been shown to be able to alleviate the inherent bias of the max-operator (Eq.~\ref{eq::act_select})~\cite{everitt2011miscellaneous}. 
Preventing bias due to the max-operator allows MPC-MFRL to operate with inaccurate forward dynamics models caused by, for example, underfitting or overfitting forward dynamics models~\cite{burnham2003model}.



\begin{algorithm}[htb]
    \caption{MPC-MFRL (Evaluation)}
    \label{alg::mbl_act}
    \begin{algorithmic}
        \STATE \textbf{Input:} a control policy $\pi_{\boldsymbol{\theta}^\pi}$, a value function $V_{\boldsymbol{\theta}^V}$, a forward dynamics model $f_{\boldsymbol{\theta}^f}$, number of simulated trajectories $N$, planning horizon $H$, number of best action sequences $E$, a task-specific reward function $R$
        \STATE \textbf{Output:} $\boldsymbol{a}^*_{t} $ 
   
        \STATE \textbf{i. Trajectory sampling}
        \STATE $\Phi \leftarrow \emptyset$
        \FOR{n $ \leftarrow 1,\dots,N$}
            \STATE $\hat{\tau}^n \leftarrow \emptyset$ 
            \STATE $\boldsymbol{\hat{s}}^n_1 \leftarrow \boldsymbol{s}_t$ 
            \FOR{h $ \leftarrow 1,\dots,H$}
                \STATE $\boldsymbol{\hat{a}}^n_h \sim \pi_{\boldsymbol{\theta}^\pi}(\boldsymbol{a}|\boldsymbol{\hat{s}}^n_h)$
                \STATE $\boldsymbol{\hat{s}}^n_{h+1} \leftarrow f_{\boldsymbol{\theta}^f}(\boldsymbol{\hat{s}}^n_h, \boldsymbol{a}^n_h)$
                \STATE $\hat{\tau}^n \leftarrow \hat{\tau}^n \cup \{\boldsymbol{\hat{s}}^n_h, \boldsymbol{a}^n_h\}$
            \ENDFOR
            \STATE $\hat{\tau}^n \leftarrow \hat{\tau}^n \cup \{\boldsymbol{\hat{s}}^n_{H+1}\}$
            \STATE $\Phi \leftarrow \Phi \cup \{\hat{\tau}^n\}$  
        \ENDFOR
     
        \STATE \textbf{ii. Trajectory evaluation}
        \STATE Compute $[\hat{G}(\boldsymbol{\hat{s}}^n_{1:H+1}, \boldsymbol{\hat{a}}^n_{1:H})]_{n=1}^{N}$ according to Eq.~\ref{eq::new_traj_eval}
      
        \STATE \textbf{iii. Action selection}
        \STATE Compute $\boldsymbol{a}^*_{t:t+H}$  according to Eq.~\ref{eq::new_act_select}
    \end{algorithmic}
\end{algorithm}

\section{Experiments}
\label{sec::exp}
The experiments are designed to answer the main question of whether MPC-MFRL successfully leverages MPC to bridge the gap between MBRL and MFRL, overcoming the drawbacks of prior works. Moreover, we perform additional experiments to answer the following more detailed questions: (1) Do we obtain a more accurate forward dynamics model by collecting training samples using an exploratory policy of MFRL? (2) Does an MFRL control policy enhance the planning performance of MPC? (3) Can the value function favor the performance of MPC planning even in an approximated forward dynamics model?
(4) Does the proposed soft-greedy action selection improve performance under forward dynamics model approximation errors?
Next, we shortly introduce the experimental setup, and then discuss the experimental results.

\subsection{Experimental setup}
\label{subsec::exp_setup}
\paragraph{Benchmark tasks.} ~
We use standard continuous control Mujoco benchmark tasks of varying difficulty from OpenAI gym~\cite{gym} 
varying in dimensions of $\mathcal{S}$ and $\mathcal{A}$: \textit{Swimmer} $(\mathcal{S} \subseteq \mathbb{R}^{10}, \mathcal{A} \subseteq \mathbb{R}^2)$, \textit{Reacher} $(\mathcal{S} \subseteq \mathbb{R}^{11}, \mathcal{A} \subseteq \mathbb{R}^2)$, 
\textit{HalfCheetah} $(\mathcal{S} \subseteq \mathbb{R}^{18}, \mathcal{A} \subseteq \mathbb{R}^6)$, 
and \textit{Ant} $(\mathcal{S} \subseteq \mathbb{R}^{30}, \mathcal{A} \subseteq \mathbb{R}^8)$. 

\paragraph{Implementation of MPC-MFRL.} ~
The MFRL control policy $\pi_{\boldsymbol{\theta^\pi}}$, the MFRL value function $V_{\boldsymbol{\theta^V}}$, and the approximated forward dynamics model $f_{\boldsymbol{\theta^f}}$ are implemented as neural networks. $\pi_{\boldsymbol{\theta^\pi}}$ is modeled as a multi-variate Gaussian distribution, where $\pi_{\boldsymbol{\theta^\pi}}(\boldsymbol{a}|\boldsymbol{s_t}) =  \mathcal{N}( {\boldsymbol{\mu_{\theta^\pi}}}, {\boldsymbol{\Sigma_{\theta^\pi}}} | \boldsymbol{s_t})$, where ${\boldsymbol{\mu_{\theta^\pi}}}$ and ${\boldsymbol{\Sigma_{\theta^\pi}}}$ are the mean vector and the covariance matrix conditioned on learned parameters $\boldsymbol{\theta^\pi}$ and current state $\boldsymbol{s}_t$. We optimize $\pi_{\boldsymbol{\theta^\pi}}$ by TRPO~\cite{schulman2015trust}, while updating $V_{\boldsymbol{\theta^V}}$ and $f_{\boldsymbol{\theta^f}}$ using the Adam optimizer~\cite{kingma2014adam}. 


\paragraph{Baselines.} ~
For comparison against both MFRL and MPC-MBRL methods we select the following baselines (see supplementary material for details):
\begin{itemize}
	\item \textit{MF (S)}: TRPO that uses stochastic actions for evaluation.
	\item \textit{MF (D)}: TRPO that uses deterministic actions for evaluation (i.e. $\boldsymbol{\mu_{\theta^\pi}}$).
	
	\item \textit{MPC-Random}: MPC that samples actions from a uniform distribution for trajectory roll-outs (Eq.~\ref{eq::traj_sampling}) and uniform random exploration with on-policy data aggregation (see Section~\ref{subsec::mpc}) for training the forward dynamics model.
	
	\item \textit{MPC-CEM}: MPC that uses the same model training approach as \textit{MPC-Random}, while using CEM~\cite{rubinstein1999cross} for trajectory sampling since CEM has been shown to work well with MPC in prior work~\cite{chua2018deep}.

\end{itemize}

\subsection{Evaluation procedure}
\label{sec:evaluation_procedure}
In order to assess performance and data-efficiency of each method, we evaluate offline~\cite{chua2018deep} each method periodically w.r.t. number of training samples used. At offline evaluation time, we fix all model parameters and measure the average total reward over 10 episodes. We then report the mean and bootstrapped confidence interval of the best average total reward (denoted "Average return" in figures below) so far over 5 distinct random seeds. 


\subsection{The results of overall performance}
\label{subsec::res_overallperf}

\begin{figure*}[htb!]
    \centering
    \begin{minipage}[b]{.99\textwidth}
    	\centering
    	\includegraphics[width=1.0\linewidth]{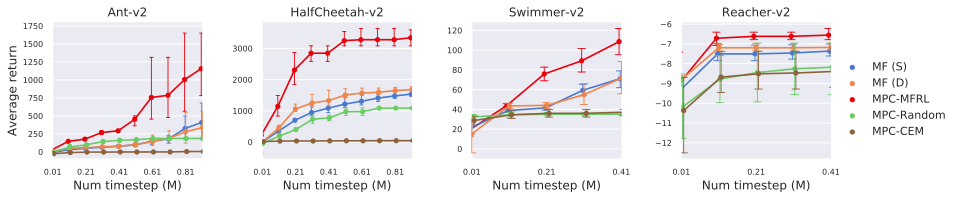}
    	\caption{Mean and bootstrapped confidence interval (solid lines and error bars, over 5 distinct random seeds) of "Average return" (see Section~\ref{sec:evaluation_procedure} for evaluation details and definition of "Average return") for different methods. "Num. timestep (M)" is the number of millions of interactions with the environment. Our method MPC-MFRL outperforms comparison methods. For comparison method and evaluation details see Section~\ref{subsec::exp_setup}.}
    	\label{fig::overall_perf}
    \end{minipage}
\end{figure*}

Fig.~\ref{fig::overall_perf} shows the performance of each method w.r.t.\ the number of samples. MPC-MFRL achieves better performance than all baseline methods: MPC-MFRL exceeds the performance of MPC-MBRL while being more data-efficient than MFRL. Moreover, the improvement is particularly significant in the more challenging tasks like \textit{Ant} and \textit{HalfCheetah}. To conclude, this result verifies the effectiveness of MPC-MFRL on common benchmark tasks.

Interestingly, Fig.~\ref{fig::overall_perf} shows that \textit{MPC-CEM} loses to \textit{MPC-Random} in \textit{Ant-v2} and \textit{HalfCheetah-v2}. Also, we find that even though \textit{MPC-CEM} obtains the highest expected return in simulation (Eq.~\ref{eq::traj_eval}), the expected return in the real environment is surprisingly low. Prior works of MBRL~\cite{sutton2012dyna,kurutach2018modelensemble} suggest that training a policy using fictitious data impairs performance. Thus, the poor performance of \textit{MPC-CEM} in complex tasks could be caused by optimizing the policy using simulated data of high approximation errors.

\subsection{The results of improved exploration for training}
\label{subsec::res_trainscheme}
\begin{figure}[htb!]
	\centering
	\subfloat[\label{fig::train_valloss}]{\includegraphics[width=0.46\linewidth]{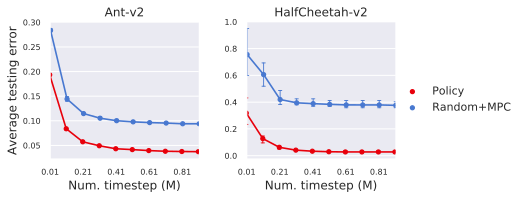}}
	\subfloat[\label{fig::train}]{\includegraphics[width=0.51\linewidth]{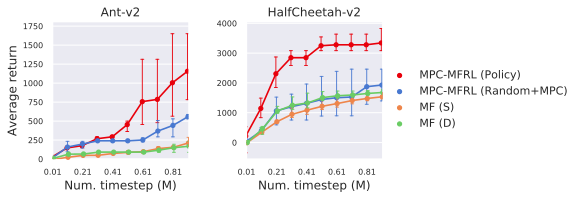}}
    \caption{(a) We measure "Average testing error" of a forward dynamics model using a pre-collected testing dataset. \textit{Policy} indicates collecting data using an MFRL policy, while \textit{Random+MPC} represents uniform random exploration with on-policy data aggregation~\cite{nagabandi2017neural}; (b) The evaluation results of MPC-MFRL with different training schemes: \textit{MPC-MFRL (Policy)} is the original MPC-MFRL, while \textit{MPC-MFRL (Random+MPC)} trains the forward dynamics model using data from \textit{Random+MPC}.The remaining legends are the same as Fig.~\ref{fig::overall_perf}. }
\end{figure}

We show that our training approach improves the training quality of forward dynamics models (Section~\ref{subsec::mpc}) by comparing various training schemes w.r.t.\ testing error of the forward dynamics model and evaluation performance.
Fig.~\ref{fig::train_valloss} shows the testing error of the forward dynamics models trained by different schemes. Half of the testing set consists of data from \textit{Random+MPC} and half from \textit{Policy} (see Section C in supplementary material for more details). \textit{Policy} reduces the testing error faster and more than \textit{Random+MPC}, which shows that collecting data by an MFRL policy can increase the accuracy of forward dynamics models. In addition to favoring accuracy, Fig.~\ref{fig::train} further shows \textit{MPC-MFRL (Policy)} outperforms \textit{MPC-MFRL (Random+MPC)}, especially in later stages of evaluation. To summarize, these results verify our approach is superior to prior approaches in terms of accuracy of forward dynamics models and evaluation performance, also showing that accuracy of forward dynamics models greatly affects the performance.

\subsection{The results of sampling trajectories by an MFRL policy}
\label{subsec::res_trajsampling}
\begin{figure}[thb!]
    \centering
    \subfloat[\label{fig::trajsample}]{\includegraphics[width=0.49\linewidth]{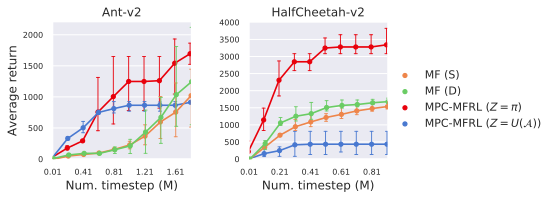}}
    \subfloat[\label{fig::trajeval}]{\includegraphics[width=0.49\linewidth]{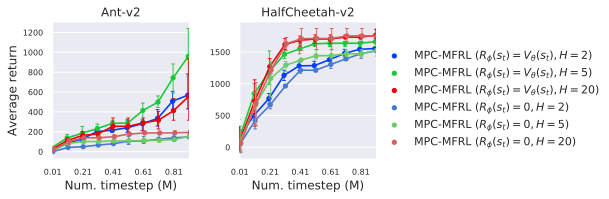}}
    \caption{(a) Varying trajectory sampling methods:  \textit{MPC-MFRL ($\mathcal{Z} = \pi$)} and \textit{MPC-MFRL ($\mathcal{Z} = U$)} respectively denote the original MPC-MFRL and MPC-MFRL that replaces the MFRL policy with an uniform distribution for trajectory sampling. See Fig.~\ref{fig::overall_perf} for details on the notation in the figure; (b) The evaluation results of different trajectory evaluation methods and planning horizons: \textit{MPC-MFRL $(R_\phi(s_t) = V(s_t), H = 2)$}, for instance, indicates MPC-MFRL with $R_\phi(s_t) = V(s_t)$ and planning horizon $H = 2$; See Fig.~\ref{fig::overall_perf} for more details on figure notation.}
\end{figure}

Fig.~\ref{fig::trajsample} shows \textit{MPC-MFRL ($\mathcal{Z} = \pi$)} outperforms \textit{MPC-MFRL ($\mathcal{Z} = U$)} in all tasks and therefore suggests that the MFRL control policy $\pi_{\boldsymbol{\theta}^\pi}$ can more readily sample trajectories of high value in the real environment than the baselines. Furthermore, \textit{MPC-MFRL ($\mathcal{Z} = \pi$)} still surpasses \textit{MPC-MFRL ($\mathcal{Z} = U$)} even in the early stages where the pure model-free variants of MPC-MFRL (i.e.\ \textit{MF (S/D)}) perform poorly compared to \textit{MPC-MFRL ($\mathcal{Z} = U$)}. This result confirms that even though the MFRL control policy $\pi_{\boldsymbol{\theta}^\pi}$ has not yet converged, our method can still enhance MPC planning performance, and also show the improvement of data-efficiency against MFRL. Note that we ignore comparison with CEM due to its poor performance in Fig.~\ref{fig::overall_perf}. 


\subsection{The strength and the limitations of trajectory evaluation with a value function}
\label{subsec::res_trajeval}

We investigate the strength and limitations of our trajectory evaluation approach in this section. 
Fig.~\ref{fig::trajeval} shows that \textit{MPC-MFRL ($R_{\phi}(s_t) = V_{\theta}(s_t), H = 2,5,20$)} outperforms \textit{MPC-MFRL ($R_{\phi}(s_t) = 0, H = 2,5,20$)} with the same planning horizon and thereby verifies that our approach (Eq.~\ref{eq::new_traj_eval}) is effective. However, on the contrary to the results in the prior work~\cite{lowrey2018plan}, Fig.~\ref{fig::trajeval} shows that ~\textit{MPC-MFRL ($R_{\phi}(s_t) = V_{\theta}(s_t), H = 5$)} approximates \textit{MPC-MFRL ($R_{\phi}(s_t) = V_{\theta}(s_t), H = 20$)}, which suggests that compounding errors in a longer planning horizon may impair the performance of value function in an approximated dynamics model. In addition, we find that in the experiments of Section~\ref{subsec::res_trajsampling}, the terminal reward (i.e. $V_{\theta}(s_t)$) in simulated trajectories of \textit{MPC-MFRL ($\mathcal{Z} = U$)} are similar than \textit{MPC-MFRL ($\mathcal{Z} = \pi$)}. This observation implies that a uniform random sampling may sample states that lead overstimation of value function in approximated dynamics model. The detail investigation are left as future works.

\subsection{The results of soft-greedy action selection}
\label{subsec::res_actselect}
\begin{figure}[tb!]
    \centering
    \subfloat[\label{fig::actselect}]{\includegraphics[width=0.48\linewidth]{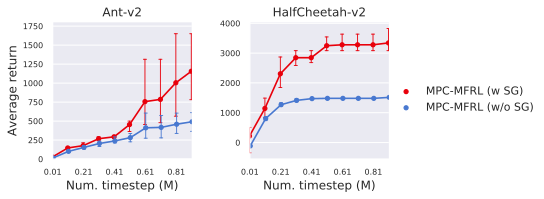}}
    \subfloat[\label{fig::sg_diff_fwd}]{\includegraphics[width=0.48\linewidth]{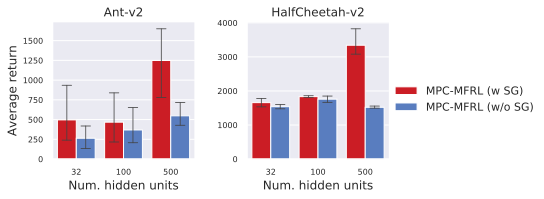}}
    \caption{(a) The evaluation results of different action selection approaches:  \textit{MPC-MFRL (w / SG)} indicates the original MPC-MFRL, while \textit{MPC-MFRL (w/o SG)} represents MPC-MFRL withoug soft-greedy action selection. The rest of legends are identical to Fig.~\ref{fig::overall_perf}; (b) Mean and bootstrapped confidence interval (bold bars and error bars, over 5 distinct random seeds) of performance for action selection approaches with different forward dynamics model complexities. "Num. hidden units" denotes the number of hidden units used. For evaluation details, see Section~\ref{sec:evaluation_procedure}}.
\end{figure}

This section verifies that our soft-greedy action selection approach improves performance under an approximate forward dynamics model, followed by studying the effectiveness with models of varying complexity.
Fig.~\ref{fig::actselect} shows that \textit{MPC-MFRL (w SG)} outperforms  \textit{MPC-MFRL (w/o SG)}, suggesting that soft-greedy action selection increases performance of an approximated forward dynamics model. Fig.~\ref{fig::sg_diff_fwd} shows that \textit{MPC-MFRL (w SG)} surpasses \textit{MPC-MFRL (w/o SG)} in all model complexities, thereby suggesting that our soft-greedy action selection is superior to classical greedy action selection in both simple and complex models. 

\section{Related work}
\label{sec::related_works}


The classic Dyna framework~\cite{sutton1990integrated,silver2008sample,kurutach2018modelensemble,kalweit2017uncertainty,clavera2018model} learn a forward dynamics model for simulating experiences to train an MFRL agent, while we focus on the interplay of MFRL and MPC.

Guided Policy Search (GPS)~\cite{levine2013guided, chebotar2017path,levine2014learning} uses model-based controllers such as iLQR~\cite{tassa2012synthesis} and iLQG~\cite{todorov2005generalized} to generate training samples for MFRL policy. The follow-up works~\cite{zhang2016learning,nagabandi2017neural} further adopt MPC to provide supervision for an policy. In contrast, our approach concentrates on improving MPC performance with MFRL. 


\citet{gu2016continuous}, \citet{feinberg2018model}, and  \citet{buckman2018sample} assist value function learning by model-based approaches. \citet{gu2016continuous} collect training samples for value function learning using a model-based controller. \citet{feinberg2018model} and \citet{buckman2018sample} leverage model-based rollout to compute targets of value function training, thus accelerating value function learning. On the contrary, our method focuses on leveraging value function to ameliorate the shortsighted planning of MPC.



\citet{silver2016mastering}, \citet{weber2017imagination}, \citet{oh2017value}, and \citet{tamar2016value} add planning to an MFRL policy. \citet{silver2016mastering} and \citet{oh2017value}, however, either rely on discrete state and action spaces or a perfect forward dynamics model. \citet{tamar2016value} assume that the state space has 2-D structure. \citet{weber2017imagination} learn planning by an MFRL approach, hence inheriting the data-inefficiency of MFRL. In contrast, our approach can be applied to an arbitrary type of state and action space and is more data-efficient than MFRL.

\citet{pong*2018temporal} perform MPC planning solely with a goal-conditioned state-action value function. \citet{lowrey2018plan} improve long-term planning by evaluating sampled trajectories using a value function.  
\citet{pong*2018temporal} and \citet{lowrey2018plan} do not utilize MFRL policies to enhance planning performance.  Planning with only a value function~\cite{pong*2018temporal} inherits the data-inefficiency of value function learning in MFRL, while our approach combines simulated rewards and a value function, thereby mitigating this problem. \citet{lowrey2018plan} assume a perfect forward dynamics model, whereas our work does not rely on such an unrealistic assumption.

\section{Conclusion}
\label{sec::conclusion}
We propose the MPC-MFRL framework which leverages the advantages of MFRL and MPC to achieve MFRL`s level of performance while being as data-efficient as MBRL. Moreover, MPC-MFRL allows the agent to continually improve performance with more environment interactions. On the other hand, our novel MPC-MFRL framework brings promising future work to light as well. An application of MPC-MFRL on real robotics systems, for example, can be an appealing direction since MPC-MFRL shows superior data-efficiency particularly crucial for real robots. Another direction could be guiding an MFRL policy by MPC. One can pre-train an MFRL policy in several tasks, then online adapt to a new task by MPC or model-based policy search. 

\bibliography{neurips_2019}
\bibliographystyle{neurips_2019}

\end{document}